\documentclass[pmlr,twocolumn,10pt]{jmlr}

\usepackage{amsmath}
\usepackage{amssymb}  
\usepackage{natbib}
\usepackage{graphicx}
\usepackage{url}
\usepackage{algorithm2e}
\usepackage{booktabs}
\usepackage{siunitx}
\usepackage{multirow}
\usepackage[most]{tcolorbox}
\usepackage{enumitem}
\usepackage{xcolor}
\usepackage{threeparttable}
\tcbset{
  colback=gray!3,
  colframe=gray!55,
  left=3mm, right=3mm, top=2mm, bottom=2mm,
  sharp corners,
  boxsep=1mm,
  breakable,
  title filled=true
}
\usepackage[switch]{lineno}

\title{LAVA: Language Model Assisted Verbal Autopsy for Cause-of-Death Determination}

\author{%
    \Name{Yiqun T. Chen} \Email{yiqunc@jhu.edu}\\
    \addr Departments of Biostatistics and Computer Science, Johns Hopkins University 
    \AND
    \Name{Tyler H. McCormick} \Email{tylermc@uw.edu} \\
        \addr Department of Statistics, University of Washington
      \AND
    \Name{Li Liu} \Email{lliu26@jhu.edu} \\
    \addr Departments of Population, Family and Reproductive Health and International Health, Johns Hopkins University 
      \AND
    \Name{Abhirup Datta} \Email{abhidatta@jhu.edu}\\
     \addr Department of Biostatistics, Johns Hopkins University 
}

\begin{document}

\maketitle

\begin{abstract}
Verbal autopsy (VA) is a critical tool for estimating causes of death in resource-limited settings where medical certification is unavailable. This study presents LA-VA, a proof-of-concept pipeline that combines Large Language Models (LLMs) with traditional algorithmic approaches and embedding-based classification for improved cause-of-death prediction. Using the Population Health Metrics Research Consortium (PHMRC) dataset across three age categories (Adult: 7,580; Child: 1,960; Neonate: 2,438), we evaluate multiple approaches: GPT-5 predictions, LCVA baseline, text embeddings, and meta-learner ensembles. Our results demonstrate that GPT-5 achieves the highest individual performance with average test site accuracies of 48.6\% (Adult), 50.5\% (Child), and 53.5\% (Neonate), outperforming traditional statistical machine learning baselines by 5-10\%. Our findings suggest that simple off-the-shelf LLM-assisted approaches could substantially improve verbal autopsy accuracy, with important implications for global health surveillance in low-resource settings.
\end{abstract}

\paragraph{Data and Code Availability}
The code and analysis scripts are publicly available at \url{https://github.com/yiqunchen/LA-VA}. The PHMRC dataset is available through the Institute for Health Metrics and Evaluation.
\section{Introduction}
Verbal autopsy (VA) is a structured interview method used to determine the probable cause of death (COD) when medical certification is unavailable~\citep{murray2014using}. This approach is especially important in low- and middle-income countries, where most deaths occur without medical attendance. Traditional VA analysis has relied on labor-intensive physician review, which is not scalable in resource-limited settings. To address this, automated algorithms such as InterVA~\citep{byass2012strengthening}, InSilicoVA~\citep{mccormick2016probabilistic}, and Tariff/smartVA~\citep{james2011performance} map symptoms to causes via probabilistic and machine-learning methods, achieving good accuracy for broad (grouped) COD categories at the individual or population level~\citep{kunihama2025bayesian, li2024bayesian,yoshida2023bayesian,datta2021regularized,fiksel2022generalized,pramanik2025modeling}.  Estimates based on these algorithms have been used to generate national and sub-national cause-specific mortality fractions, \citep[see, e.g., estimates of CSMF for children and neonates in Mozambique,][]{macicame2023countrywide}. 

Despite this progress, several methodological challenges remain. First of all, the label space can be large (hundreds of cause categories at the finest granularity), and many less-prevalent causes are data-sparse, limiting reliability. Secondly, most algorithms assume a single training dataset or knowledge base defining symptom–cause relationships, yet in real data both the marginal distribution of causes and the conditional distribution of causes given symptoms vary across settings (e.g., regions, instruments, or even data-collection teams). Moreover, baseline symptom distributions and reporting patterns also differ, producing well-documented performance heterogeneity across geographies~\citep{datta2021regularized,fiksel2022generalized,li2024bayesian,murray2014using}. Third, the interview is onerous and requires the question of surviving caregivers or family members to complete an extensive, taxing survey, essentially translating the circumstances of the death into the computer's language to apply standard ML algorithms. Finally, though VA is inherently narrative, free text data has only recently been explored as a source of information for ascertaining causes~\citep{chu2025leveraging,manaka2022improving,blanco2020extracting}, with previous approaches primarily utilizing the tabular data form containing binarized responses to the questions in the VA tool. 

The advances of Large language models (LLMs) offer several new avenues for addressing these gaps. First, prompts to LLMs can be crafted to \emph{explicitly encode distributional shifts} in how symptoms are expressed across regions. For example, one might include contextual instructions such as: \emph{``In Region A, chest pain is described as severe discomfort; in Region B, it is reported as difficulty breathing.''} By making such variation explicit, LLMs can adapt to regional heterogeneity in symptom reporting without the need for retraining on large, site-specific datasets. Second, LLMs provide a flexible platform for experimenting with different levels of cause (label) granularity. Rather than being restricted to a fixed set of cause categories, LLMs can be directed to operate at varying resolutions---for instance, treating all cardiovascular deaths as a single category or distinguishing more specific conditions such as myocardial infarction, stroke, and heart failure. In addition, their ability to generate reasoning steps and uncertainty-aware explanations offers a novel complement to existing probabilistic models~\citep{guo2025deepseek}. Finally, LLMs can effectively process free-text responses in a semantically rich manner, including in the medical setting~\citep{singhal2025toward}. This capacity allows them to capture subtle but clinically meaningful details embedded in interview transcripts that often escape structured symptom lists and keyword-based features. Recent studies demonstrate that leveraging this narrative information can significantly improve model performance on related medical reasoning tasks~\citep{manaka2024multi}. 

Motivated by these opportunities, we introduce LA-VA (\textbf{L}anguage model \textbf{A}ssisted \textbf{V}erbal \textbf{A}utopsy), an end-to-end pipeline for testing LLMs and machine-learning methods in COD prediction. We focus on three research questions (RQs):
\begin{enumerate}
  \setlength{\itemsep}{0pt}      
  \setlength{\parskip}{0pt}      
  \setlength{\topsep}{0pt}       
  \item \textbf{RQ1:} Can modern LLMs achieve better accuracy than existing VA algorithms?
  \item \textbf{RQ2:} How do different approaches (LLMs, embeddings, and traditional algorithms) complement one another in an ensemble framework?
  \item \textbf{RQ3:} What are the cause-specific and geographic patterns of performance that inform practical deployment strategies?
\end{enumerate}
\subsection{Dataset}
We analyzed a subset of the Population Health Metrics Research Consortium (PHMRC) reference standard verbal autopsy (VA) dataset~\citep{murray2011phmrc}, comprising 11{,}978 cases with physician-certified causes of death. PHMRC spans six sites in four countries — Mexico City (Mexico); Dar es Salaam and Pemba (Tanzania); Bohol (Philippines); and Andhra Pradesh and Uttar Pradesh (India) — and was designed for method validation with strict clinical criteria and standardized VA instruments. We follow the established age modules for COD classification: {Adults} (\(N{=}7{,}580\)) with 34 causes; {Children} (\(N{=}1{,}960\)) with 21 causes; and {Neonates} (\(N{=}2{,}438\)) with 6 causes. Each record includes demographics, symptom checklists, and a free-text narrative of varying length and transcription quality. The multi-country design supports evaluation under distribution shift across cultural and health-system contexts~\citep{murray2011metrics}.

\subsection{LA–VA Pipeline}
Our \emph{Language-Assisted Verbal Autopsy} (LA–VA) pipeline integrates four complementary components:

\paragraph{(1) Large-language-model (LLM) predictions.}
We use GPT-5, a state-of-the-art LLM (as of August 2025), with age-specific system prompts for direct cause prediction. Each case is formatted as a structured prompt including (i) demographics (age/sex/site), (ii) symptom responses, (iii) the narrative, and (iv) brief care-access context. Prompts guide stepwise diagnostic reasoning (differentials $\rightarrow$ evidence synthesis $\rightarrow$ ranked causes), with detailed templates provided in the Appendix.

\paragraph{(2) LCVA baseline.}
LCVA, a Bayesian latent class model designed specifically to address domain shift issues in VA data, has recently been benchmarked as a high-performing probabilistic model on structured VA data~\citep{li2024bayesian}. At a high level, it jointly models symptom patterns, accommodates cross-site heterogeneity, and produces per-case posterior cause probabilities for COD.

\paragraph{(3) Embedding-based classification.}
We embed the concatenated structured responses and narratives using \emph{voyage-3-large} embeddings ($d=1{,}024$)~\citep{voyage_docs}, and train off-the-shelf tabular models including $\ell_2$-regularized logistic regression classifiers and random forests for each cause using \texttt{scikit-learn}. This captures semantic similarity in narratives while remaining computationally efficient; we present logistic regression results in the main text.

\paragraph{(4) Meta-learner ensemble.}
We \emph{stack} the above models with a meta-learner that ingests per-method predicted probabilities and outputs final calibrated probabilities over causes. We experimented with two ensembling variants: (i) a {simple weighted average over predictions (searched over a grid); and (ii) a classifier-based approach (e.g., XGBoost or logistic regression) trained on all posteriors.

\subsection{Evaluation and Metrics}
\label{sec:evaluation}

\paragraph{Cross-site design.}
To assess geographic generalizability under realistic distribution shift, we use leave-one-site-out (LOSO) cross-validation: in each iteration, one site is held out for testing and the remaining five are used for training, mirroring deployment settings where labeled data from the target site are unavailable~\citep{li2024bayesian}. Within each fold, hyperparameters are tuned by nested CV. For stacking ensembles, we generate out-of-fold base-model probabilities (5-fold stratified CV within the training sites) to train the meta-learner, then refit base models on full in-fold data before evaluation on the held-out site. For completeness, conventional random splits (stratified by age) are reported in the Appendix, though they are less sensitive to domain shift than LOSO.

\paragraph{Individual-level metrics.}
We report Top-1 accuracy, the fraction of cases where the top prediction matches the true cause, and Top-5 accuracy, the fraction where the true cause appears anywhere among the five highest-ranked predictions.

\paragraph{Population-level metric.}
Cause-specific mortality fraction (CSMF) accuracy evaluates aggregate estimation. Let the true CSMF vector be $\boldsymbol{\pi}^{\text{true}}$ with $\pi^{\text{true}}_c = \tfrac{1}{N}\sum_i \mathbb{I}\{y_i=c\}$ and the predicted CSMF be $\hat{\pi}_c = \tfrac{1}{N}\sum_i \hat{p}_{ic}$. Following~\citet{murray2011metrics},
\begin{equation}
\label{eq:csmfacc-tight}
\mathrm{CSMF} \;=\; 1 - 
\frac{\sum_{c=1}^C \lvert \hat{\pi}_c - \pi^{\text{true}}_c \rvert }{2\,(1 - \min_c \pi^{\text{true}}_c)}.
\end{equation}
This directly measures closeness of estimated cause fractions to the truth, which is critical for public-health decision-making even when individual-level predictions remain uncertain.
\subsection{Calibrating LLM Predictions}
\label{sec:popcalib}

Reliable prediction and calibration for LLM-based classification remains an open challenge~\citep{tian2023just}, particularly with many classes. To balance predictive richness with reliability, we restrict the model to output up to five top COD predictions with confidence levels (high/medium/low), followed by lightweight post-hoc calibration that reweights GPT-5 probabilities to improve population-level metrics such as CSMF in \eqref{eq:csmfacc-tight}.  

Let $\hat{\mathbf{p}}_i$ be the predicted probability vector for case $i$ over $C$ unique CODs, and let $\pi_c$ denote the empirical prevalence of cause $c$ in the training data. We introduce nonnegative weights $\alpha_1 \ge \alpha_2 \ge \cdots \ge \alpha_5$ for the model’s top-5 predictions (generalizable to top-$N$), with $\sum_{j=1}^5 \alpha_j \le 1$. Remaining probability mass is distributed across other causes in proportion to $\pi_c$.

Formally, for top-5 predictions $\hat{\mathcal{R}}_i^{(5)}$ of case $i$, the calibrated probabilities are
\[
q_{ic}(\boldsymbol{\alpha}) =
\begin{cases}
\alpha_j, & \text{if } c = \hat{r}_{ij}, \; j \in \{1,\ldots,5\}, \\[6pt]
\bigl(1 - \sum_{j=1}^5 \alpha_j \bigr)\,\tilde{\pi}_c, & \text{if } c \notin \hat{\mathcal{R}}_i^{(5)},
\end{cases}
\]
with normalized prevalence
$
\tilde{\pi}_c = {\pi_c}/{\sum_{c' \notin \hat{\mathcal{R}}_i^{(5)}} \pi_{c'}}.
$

We estimate $\boldsymbol{\alpha}$ by minimizing the discrepancy between the calibrated distribution $\bar{q}_c = \tfrac{1}{N}\sum_i q_{ic}$ and the training distribution $r_c$:
$
\min_{\boldsymbol{\alpha} \in \mathbb{R}_{\ge 0}^5}
\sum_{c=1}^C \bigl| \bar{q}_c(\boldsymbol{\alpha}) - r_c \bigr|,
\text{s.t. } \sum_{j=1}^5 \alpha_j \le 1.
$
This linear program can be solved efficiently with convex optimization packages~\citep{diamond2016cvxpy}. We allow distinct calibration parameters for different confidence levels.

\section{Results}
\paragraph{Overall Performance}

\begin{figure*}[t]
\centering
\includegraphics[width=1.02\linewidth]{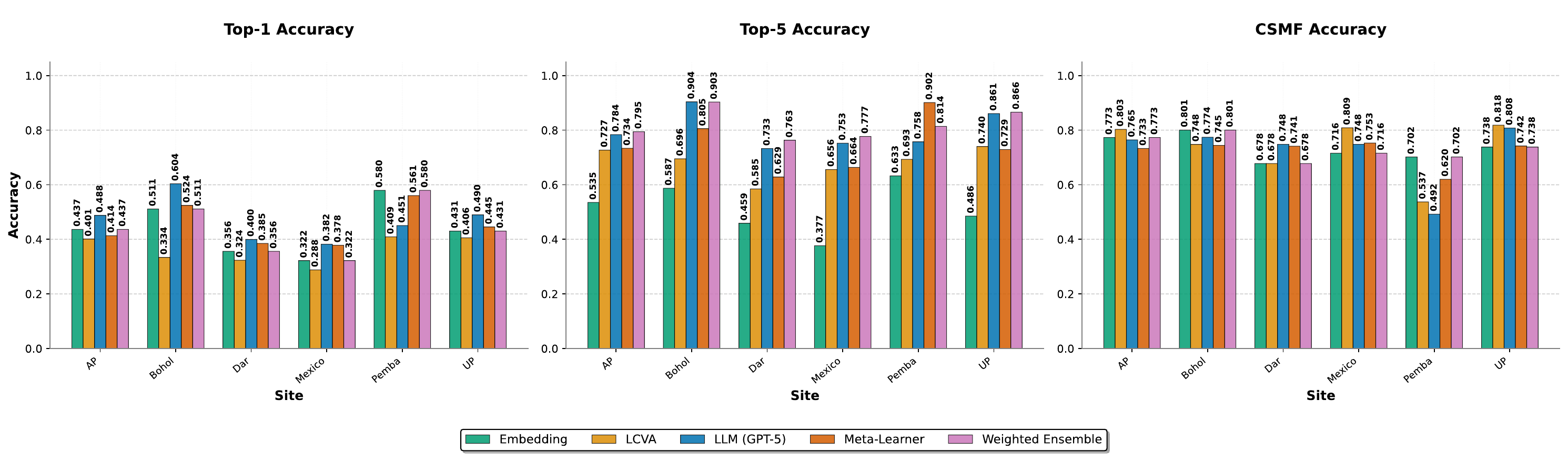}
\vspace{-2em}
\caption{Overall performance of all methods for adult data on Top-1, Top-5, and CSMF accuracy.}
\label{fig:adult_performance}
\vspace{-1.5em}
\end{figure*}
Figure~\ref{fig:adult_performance} summarizes performance for the adult data, the subset with the most benchmark results~\citep{li2024bayesian}. GPT-5 consistently achieved the highest accuracy among individual models, surpassing both classical baselines (e.g., LCVA) and embedding-based classifiers. Gains were evident in both individual-level metrics (Top-1/Top-5) and, to a less extent, population-level estimation (CSMF accuracy). As in prior work with probabilistic models, we observed substantial site-level heterogeneity. For example, in Pemba, embedding-based classifiers outperformed both GPT-5 and LCVA, whereas in most other sites, GPT-5 was dominant. Overall, there was little difference between the meta-learner and the weighted ensemble.  

We note that raw GPT-5 predictions alone did not consistently outperform probabilistic Bayesian methods such as LCVA at the population level when measured by CSMF. To address this, we applied the calibration procedure described in Section~\ref{sec:popcalib} to GPT-5 predictions using the training set, aiming to improve population-level assessment. Table~\ref{tab:loso_detailed_comparison} reports these calibrated results, including an additional column for calibrated GPT-5. Calibration yielded consistent gains in CSMF accuracy, raising the adult mean by 4\% to 77\%, compared to 0.71 for LCVA. Importantly, calibration by design has \emph{no impact} on Top-$N$ accuracy, underscoring its value as a lightweight adjustment for better population-level estimations.
\begin{table}[htbp!]
\centering
\caption{CSMF Accuracy for Adults by Site; best in bold and \underline{second best} underlined.}
\label{tab:loso_detailed_comparison}
\footnotesize
\begin{tabular}{@{}llccc@{}}
\toprule
Age Group & Site & GPT\textsubscript{Orig} & GPT\textsubscript{Cali} & LCVA \\
\midrule
\multirow{6}{*}{Adult} 
 & AP     & 0.78 & \textbf{0.85} & 0.80 \\
 & Bohol  & \underline{0.79} & \textbf{0.80} & 0.71 \\
 & Dar    & \underline{0.77} & \textbf{0.81} & 0.70 \\
 & Mexico & 0.71 & \textbf{0.83} & \underline{0.75} \\
 & Pemba  & \underline{0.58} & 0.53 & 0.46 \\
 & UP     & 0.76 & \underline{0.80} & \textbf{0.82} \\
\midrule
\multicolumn{2}{l}{\textbf{Adult Mean}} & 0.73 (0.1) & \textbf{0.77} (0.1) & 0.71 (0.1) \\
\bottomrule
\end{tabular}
\end{table}

\paragraph{Cause-Specific Performance}
Performance for adults varied widely across causes. The highest accuracies were observed for conditions with distinctive and easily recognizable presentations: maternal deaths were identified with over 90\% accuracy and breast cancer with $\sim$85\%. External causes such as road traffic accidents (73\%) were also captured reliably, reflecting that traumatic or context-specific events are often described unambiguously in narratives. By contrast, accuracy was substantially lower for conditions with overlapping or nonspecific symptoms. Cardiovascular diseases such as myocardial infarction and stroke were frequently confused (see Appendix~\ref{app:cause_specific_accuracy} for full cause accuracy distribution).

\paragraph{Narrative Length and Accuracy.} Narratives were present in 87.1\% of cases (9,113/10,466), with a mean length of $368.5 \pm 386.8$ characters (median 234; range 2–2,435). Accuracy generally rose with length: 49.4\% ($<$250), 54.1\% (251–500), 58.0\% (501–1,000), and 63.2\% ($>$1,000; $p<0.001$). 

\section{Discussion and Conclusion}
LLM-assisted verbal autopsy substantially improves cause-of-death classification, achieving 48--54\% top-1 accuracy across age groups and exceeding prior state-of-the-art results by 5--10\%. Gains were largest for causes with distinctive presentations, such as maternal mortality and homicide, underscoring the potential of LLMs for tracking unambiguous outcomes. More ambiguous conditions, including cardiovascular subtypes and certain cancers, remained challenging, motivating further work on in-context learning and fine-tuning strategies~\citep{ziegler2019fine}. Post-training calibration with limited target-domain data, previously shown to improve population-level accuracy for statistical VA algorithms \citep{datta2021regularized,fiksel2023correcting,gilbert2023multi}, may also benefit LLM-based approaches.  

Key limitations include reliance on interview quality, use of historical PHMRC data (2007--2010), and computational and privacy concerns that may restrict deployment in resource-limited or identifiable settings. Because open-source narratives are almost exclusively in English, broader multilingual validation is essential, including efforts with multimodal VA datasets such as COMSA~\citep{macicame2023countrywide} and CHAMPS~\citep{farag2017precisely}.  

Overall, LA-VA shows that LLMs can advance mortality surveillance where medical certification is lacking. Future directions include integrating complementary data and ensuring reproducible deployment through open-weight models.

\bibliography{references}

\clearpage
\appendix
\onecolumn  

\section{Additional Details on the LLM Prompts}
\label{sec:prompt_example}

Verbal autopsy (VA) narratives present unique challenges for automated cause-of-death
(COD) assignment. Unlike structured clinical data, VA responses contain free-text
descriptions from lay interviewers, variable symptom checklists, and contextual
details (e.g., access to care, location of death). These data require interpretation
in light of both epidemiological priors (what conditions are common in a given age
group or setting) and fine-grained clinical heuristics (how to distinguish similar
syndromes such as pneumonia vs.\ sepsis, or stroke vs.\ myocardial infarction).

To encode such reasoning within large language models (LLMs), we developed a family
of structured, age-specific system prompts. Each template balances three goals:

\begin{enumerate}[noitemsep,topsep=2pt]
    \item \textbf{Restrict outputs to valid COD categories.} Each age group has a
    pre-specified label set (34 adult, 21 child, 6 neonatal), derived from the
    PHMRC reference standard verbal autopsy study and WHO cause lists.
    \item \textbf{Encourage consistent diagnostic reasoning.} Prompts specify a
    causal hierarchy (immediate $\rightarrow$ underlying $\rightarrow$ contributing
    factors), a temporal framework (onset $\rightarrow$ progression $\rightarrow$ death),
    and explicit differential diagnosis criteria.
    \item \textbf{Promote transparency and reproducibility.} Each output includes
    not only a COD label but also a one-sentence rationale citing the temporal
    sequence, immediate mechanism, and comorbidity weighting.
\end{enumerate}

We provide the overall system prompt below (Appendix~\ref{app:base-prompt}), followed
by age-specific guidance blocks and examples. The placeholders
\texttt{\{age\_group\}}, \texttt{\{cod\_list\}}, and \texttt{\{examples\}} are
instantiated differently for adults, children, and neonates.

\subsection{Base System Prompt Template}
\label{app:base-prompt}

\begin{tcolorbox}[title=Base System Prompt, width=\textwidth,
  colback=gray!3, colframe=gray!55,
  left=3mm, right=3mm, top=2mm, bottom=2mm,
  sharp corners, boxsep=1mm, breakable]
\small\ttfamily
You are an experienced verbal-autopsy coder with medical expertise specializing in \{age\_group\} deaths.

\medskip
TASK
1. Think step-by-step (hide the chain-of-thought).  
2. Output \emph{one} cause label from the allowed list.  
3. Provide a single-sentence rationale citing:  
   (a) key temporal clues,  
   (b) the immediate mechanism, and  
   (c) how comorbidities/injuries were weighed.  

\medskip
ALLOWED CAUSES FOR \{age\_group\_upper\}:  
\{cod\_list\}  

\medskip
ANALYSIS FRAMEWORK  
-- Build an internal timeline: onset $\rightarrow$ progression $\rightarrow$ death  
-- Distinguish the primary underlying cause from complications  
-- Separate similar conditions using explicit criteria  
-- For external causes: assess intentionality (accident vs.\ intentional harm)  
-- Balance epidemiological context with individual case specifics  
-- Consider age-specific patterns and vulnerabilities  

\medskip
\{age\_specific\_guidance\}  

\medskip
PRIMARY CAUSE HIERARCHY  
1. Immediate mechanism (what acutely stopped heart/breathing)  
2. Underlying disease (what led to that mechanism)  
3. Contributing factors (what worsened the course)  

\medskip
\{examples\}
\end{tcolorbox}

\subsection{Instantiation Across Age Groups}
For each age group, the template is instantiated by: (i) inserting the appropriate COD label set; (ii) injecting the corresponding diagnostic guidance; and (iii) appending representative few-shot exemplars. Below we show concrete instantiations (abridged guidance excerpts; full guidance appears in Appendix~Z).

\subsubsection*{Adult (Instantiated Example)}
\begin{tcolorbox}[title=Adult System Prompt (instantiated), width=\textwidth,
  colback=gray!3, colframe=gray!55, left=3mm, right=3mm, top=2mm, bottom=2mm,
  sharp corners, boxsep=1mm, breakable]
\small\ttfamily
You are an experienced verbal-autopsy coder with medical expertise specializing in adult deaths.

\medskip
TASK
1. Think step-by-step (hide the chain-of-thought).  
2. Output \emph{one} cause label from the allowed list.  
3. Provide a single-sentence rationale citing temporal clues, immediate mechanism, and comorbidity weighting.

\medskip
ALLOWED CAUSES FOR ADULTS:  
AIDS; Acute Myocardial Infarction; Asthma; Bite of Venomous Animal; Breast Cancer; COPD; Cervical Cancer; Cirrhosis; Colorectal Cancer; Diabetes; Diarrhea/Dysentery; Drowning; Epilepsy; Esophageal Cancer; Falls; Fires; Homicide; Leukemia/Lymphomas; Lung Cancer; Malaria; Maternal; Other Cardiovascular Diseases; Other Infectious Diseases; Other Injuries; Other Non-communicable Diseases; Pneumonia; Poisonings; Prostate Cancer; Renal Failure; Road Traffic; Stomach Cancer; Stroke; Suicide; TB.

\medskip
ANALYSIS FRAMEWORK  
-- Timeline: onset $\rightarrow$ progression $\rightarrow$ death  
-- Primary vs.\ complications; explicit differentials; intentionality for external causes; age-patterns.

\medskip
\textit{Adult guidance (excerpt).} OTHER CVD vs.\ ACUTE MI: MI $<$1h onset with crushing chest pain and collapse; OTHER CVD shows chronic dyspnea/edema with gradual decline. AIDS vs.\ TB: AIDS with chronic wasting/opportunistic infections; TB with cough $>$2~weeks, night sweats, hemoptysis.

\medskip
PRIMARY CAUSE HIERARCHY  
1. Immediate mechanism \quad 2. Underlying disease \quad 3. Contributing factors

\medskip
\# -------- FEW-SHOT EXAMPLE --------\\
\textbf{DEMOGRAPHICS:} 55-y male, Mexico\\
\textbf{QUESTIONNAIRE:} sudden chest pain $<$1 h, no fever/cough\\
\textbf{NARRATIVE:} ``He clutched his chest while working and collapsed…''\\
\textbf{Cause:} Acute Myocardial Infarction\\
\textbf{Rationale:} Abrupt chest pain + hyperacute collapse indicates AMI; diabetes noted but not terminal.\\
\# -------- END EXAMPLE --------
\end{tcolorbox}

\subsubsection*{Child (Instantiated Example)}
\begin{tcolorbox}[title=Child System Prompt (instantiated), width=\textwidth,
  colback=gray!3, colframe=gray!55, left=3mm, right=3mm, top=2mm, bottom=2mm,
  sharp corners, boxsep=1mm, breakable]
\small\ttfamily
You are an experienced verbal-autopsy coder with medical expertise specializing in child deaths.

\medskip
TASK
1. Think step-by-step (hide the chain-of-thought).  
2. Output \emph{one} cause label from the allowed list.  
3. Provide a single-sentence rationale citing temporal clues, immediate mechanism, and comorbidity weighting.

\medskip
ALLOWED CAUSES FOR CHILDREN:  
AIDS; Bite of Venomous Animal; Diarrhea/Dysentery; Drowning; Encephalitis; Falls; Fires; Hemorrhagic fever; Malaria; Measles; Meningitis; Other Cancers; Other Cardiovascular Diseases; Other Defined Causes of Child Deaths; Other Digestive Diseases; Other Infectious Diseases; Pneumonia; Poisonings; Road Traffic; Sepsis; Violent Death.

\medskip
ANALYSIS FRAMEWORK  
-- Timeline and causal hierarchy; explicit pediatric differentials; vaccination/nutrition context; environmental hazards.

\medskip
\textit{Child guidance (excerpt).} PNEUMONIA vs.\ SEPSIS: Pneumonia with cough, fever, fast breathing (2--7~days); Sepsis with rapid multi-organ decline, poor feeding, lethargy.

\medskip
PRIMARY CAUSE HIERARCHY  
1. Immediate mechanism \quad 2. Underlying disease \quad 3. Contributing factors

\medskip
\# -------- FEW-SHOT EXAMPLE --------\\
\textbf{DEMOGRAPHICS:} 3-y male, rural Tanzania\\
\textbf{QUESTIONNAIRE:} fever 3 days, convulsions, unconscious\\
\textbf{NARRATIVE:} ``High fever that comes and goes, then had fits and stopped responding''\\
\textbf{Cause:} Malaria\\
\textbf{Rationale:} Cyclical fever with seizures and coma in an endemic area indicates cerebral malaria.\\
\# -------- END EXAMPLE --------
\end{tcolorbox}

\subsubsection*{Neonate (Instantiated Example)}
\begin{tcolorbox}[title=Neonate System Prompt (instantiated), width=\textwidth,
  colback=gray!3, colframe=gray!55, left=3mm, right=3mm, top=2mm, bottom=2mm,
  sharp corners, boxsep=1mm, breakable]
\small\ttfamily
You are an experienced verbal-autopsy coder with medical expertise specializing in neonate deaths.

\medskip
TASK
1. Think step-by-step (hide the chain-of-thought).  
2. Output \emph{one} cause label from the allowed list.  
3. Provide a single-sentence rationale citing temporal clues, immediate mechanism, and comorbidity weighting.

\medskip
ALLOWED CAUSES FOR NEONATES:  
Birth asphyxia; Congenital malformation; Meningitis/Sepsis; Pneumonia; Preterm Delivery; Stillbirth.

\medskip
ANALYSIS FRAMEWORK  
-- Emphasize timing relative to birth ($<$24h, 1--7d, $>$7d); maternal/delivery factors; birth weight/gestation.

\medskip
\textit{Neonate guidance (excerpt).} BIRTH ASPHYXIA vs.\ PRETERM: Asphyxia with failure to cry/breathe at birth and early seizures; Preterm with very low birth weight, persistent respiratory distress, temperature instability.

\medskip
PRIMARY CAUSE HIERARCHY  
1. Immediate mechanism \quad 2. Underlying disease \quad 3. Contributing factors

\medskip
\# -------- FEW-SHOT EXAMPLE --------\\
\textbf{DEMOGRAPHICS:} 2-day male, home delivery\\
\textbf{QUESTIONNAIRE:} didn't cry at birth, difficulty breathing\\
\textbf{NARRATIVE:} ``Long labor, baby didn't cry when born, tried to make him breathe''\\
\textbf{Cause:} Birth asphyxia\\
\textbf{Rationale:} Failure to establish breathing at birth after prolonged labor indicates asphyxia.\\
\# -------- END EXAMPLE --------
\end{tcolorbox}

\paragraph{Forward-looking.}
We plan to integrate these handcrafted templates with emerging \emph{autoprompting} and prompt-optimization methods (e.g., DsPy~\citep{khattab2024dspy}) to adapt instructions automatically and further improve robustness and accuracy across sites and age modules.

\section{Additional Results for Other Age Groups}

For children, overall performance was higher than adults in relative terms, though more variable across sites. GPT-5 achieved top-1 accuracies of 30--60\% and top-5 above 70\%, with CSMF accuracies up to 0.77 in Bohol. Calibration provided small but consistent gains. Embedding-based classifiers and ensembles showed similar trends, though encoding choices for the child module appear suboptimal. Importantly, independent baselines are scarce: most traditional probabilistic models require pediatric-specific modifications and have not been benchmarked separately. This gap underscores the practical advantage of LLM-based approaches, which transfer across modules without additional engineering.

For neonates, performance was stronger still. GPT-5 top-1 accuracies ranged from 60--80\%, with CSMF reaching 0.92 in Dar. Because there are only six possible causes of death, top-5 accuracy is not meaningful and is therefore omitted. LCVA and related probabilistic models do not currently support neonates, again highlighting the ease of extension for LLMs. Calibration improved aggregate accuracy further without altering individual-level predictions.

Together, these results show that while benchmark baselines exist primarily for adults, LLMs extend naturally to children and neonates. Their ability to operate without module-specific adjustments represents a key strength for deployment in diverse real-world settings where traditional methods face practical limitations.

\begin{figure}[htbp!]
    \centering
    \includegraphics[width=\linewidth]{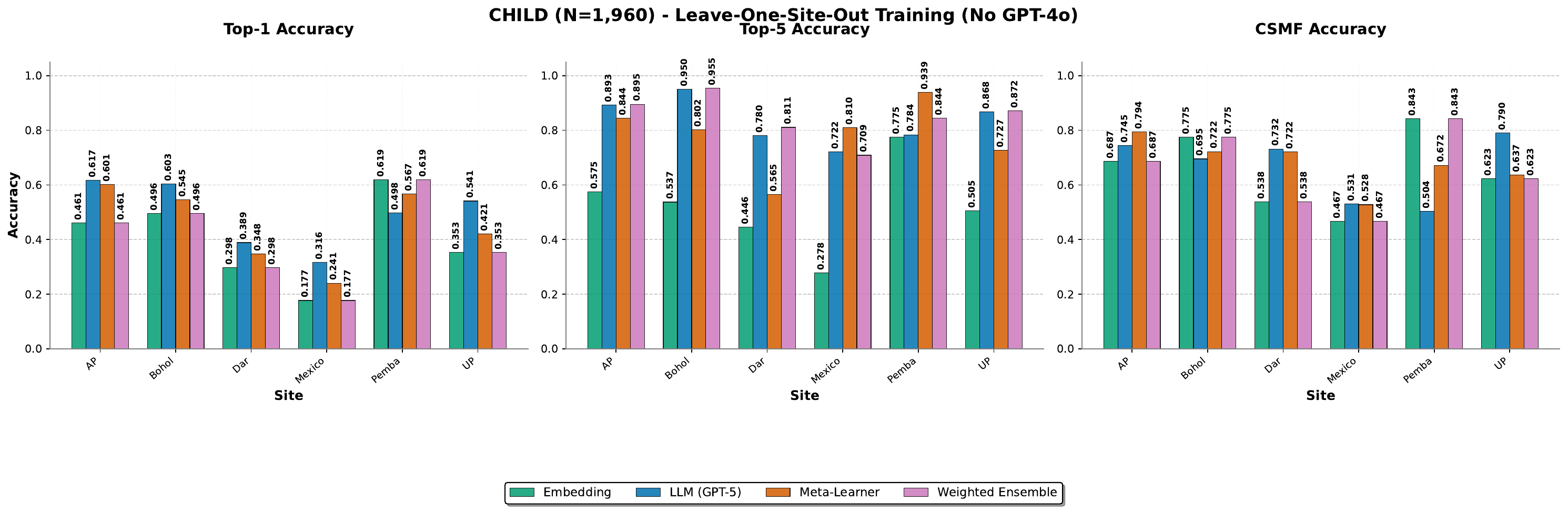}
    \caption{Child performance across sites.}
    \label{fig:child_results}
\end{figure}

\begin{figure}[htbp!]
    \centering
    \includegraphics[width=\linewidth]{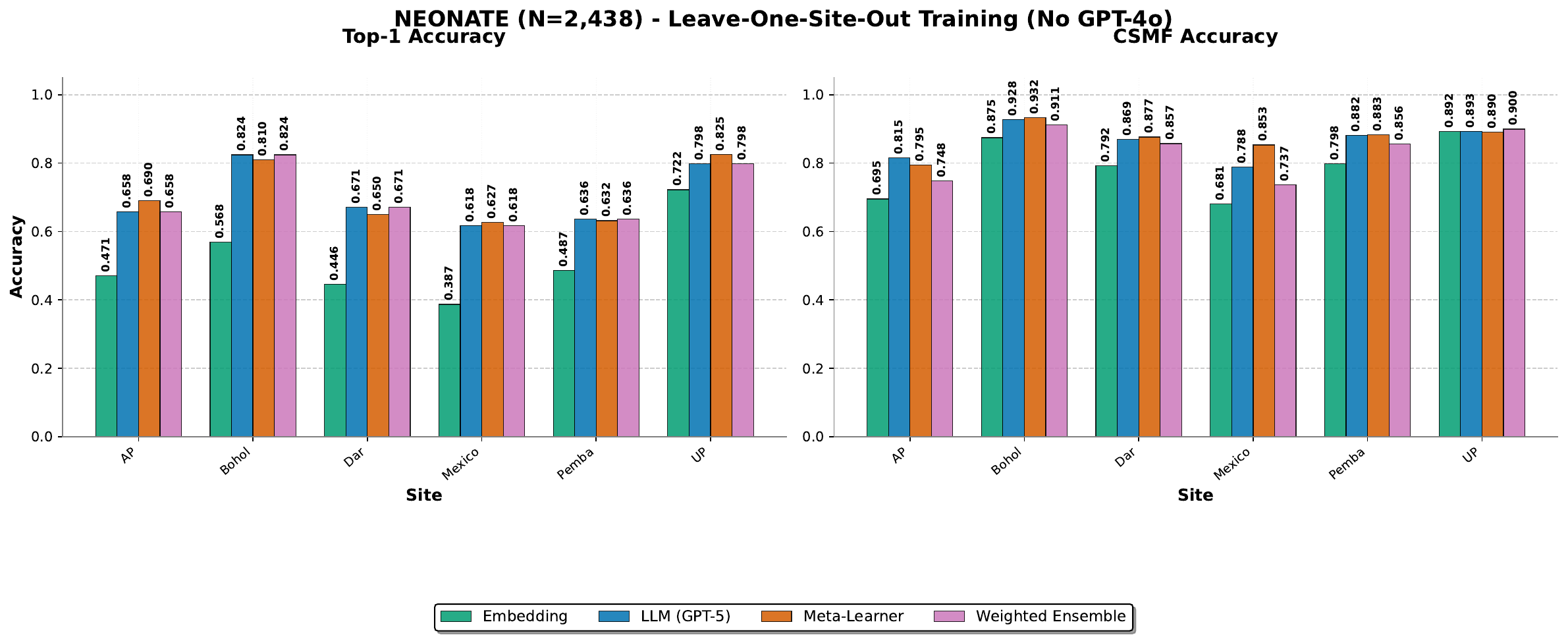}
    \caption{Neonate performance across sites.}
    \label{fig:neonate_results}
\end{figure}

\section{Cause-Specific Results}
\label{app:cause_specific_accuracy}

For adults, accuracy is highest for distinctive causes (e.g., Road Traffic, Maternal, Stroke, TB, Cirrhosis), and lower for heterogeneous categories (Other Infectious/Non-communicable). Sites with richer narratives (e.g., Bohol) generally show stronger performance than those with terse narratives (e.g., Dar).

\begin{table}[htbp!]
\centering
\caption{Cause-specific Top-1 accuracy (Adult) by site, with narrative context.}
\label{tab:cause_specific_adult}
\begin{threeparttable}
\begin{tabular}{lcccccc}
\toprule
\textbf{Cause of Death} & \textbf{AP} & \textbf{Bohl} & \textbf{Dar} & \textbf{Mex} & \textbf{Pmb} & \textbf{UP} \\
\midrule
\textit{Narratives \%} & \textit{74} & \textit{96} & \textit{100} & \textit{88} & \textit{98} & \textit{97} \\
\textit{Avg Words}     & \textit{90} & \textit{175} & \textit{12}  & \textit{63} & \textit{16} & \textit{57} \\
\midrule
Pneumonia                 & 0.38 & 0.29 & 0.12 & 0.23 & 0.14 & 0.13 \\
Stroke                    & 0.70 & 0.82 & 0.65 & 0.50 & --   & 0.84 \\
Other Non-communicable    & 0.16 & 0.29 & 0.10 & 0.06 & 0.17 & 0.03 \\
AIDS                      & 0.44 & --   & 0.53 & 0.65 & --   & 0.72 \\
Other Cardiovascular      & 0.32 & 0.59 & 0.15 & 0.36 & --   & 0.32 \\
Maternal                  & 0.90 & 0.93 & 0.83 & 0.79 & 0.65 & 0.94 \\
Diarrhea/Dysentery        & 0.31 & 0.58 & 0.19 & 0.06 & 0.37 & 0.56 \\
Renal Failure             & 0.46 & 0.72 & 0.39 & 0.36 & --   & 0.44 \\
Diabetes                  & 0.34 & 0.40 & 0.19 & 0.24 & 0.45 & 0.06 \\
Acute Myocardial Infarct. & 0.43 & 0.22 & 0.00 & 0.18 & --   & 0.65 \\
Other Infectious          & 0.24 & 0.63 & 0.08 & 0.09 & 0.67 & 0.26 \\
Cirrhosis                 & 0.71 & 0.87 & 0.30 & 0.78 & --   & 0.62 \\
Road Traffic              & 0.87 & 0.94 & 0.81 & 0.39 & 0.78 & 0.44 \\
TB                        & 0.81 & 0.73 & 0.35 & 0.65 & 0.83 & 0.67 \\
Malaria                   & 0.21 & --   & 0.26 & --   & 0.00 & 0.26 \\
Falls                     & 0.44 & 0.76 & 0.27 & 0.26 & 0.64 & 0.27 \\
Drowning                  & 0.59 & 0.75 & 0.78 & --   & 0.75 & 0.15 \\
Fires                     & 0.54 & 0.83 & 0.42 & 0.25 & 0.50 & 0.13 \\
Breast Cancer             & 0.33 & 1.00 & 0.91 & 0.75 & --   & 1.00 \\
\bottomrule
\end{tabular}
\begin{tablenotes}\small
\item ``Bohl''=Bohol, ``Mex''=Mexico, ``Pmb''=Pemba. ``--'' indicates no cases for that site/cause. Top rows summarize narrative availability and length.
\end{tablenotes}
\end{threeparttable}
\end{table}

For children, pneumonia and diarrheal diseases are reasonably well identified, while performance varies more sharply for injuries (falls, drowning) by site. Narrative richness (e.g., Bohol) again aligns with higher accuracy.
\begin{table}[ht]
\centering
\caption{Cause-specific Top-1 accuracy (Child) by site, with narrative context.}
\label{tab:cause_specific_child}
\begin{threeparttable}
\begin{tabular}{lcccccc}
\toprule
\textbf{Cause of Death} & \textbf{AP} & \textbf{Bohl} & \textbf{Dar} & \textbf{Mex} & \textbf{Pmb} & \textbf{UP} \\
\midrule
\textit{Narratives \%} & \textit{59} & \textit{98} & \textit{100} & \textit{72} & \textit{98} & \textit{92} \\
\textit{Avg Words}     & \textit{102} & \textit{184} & \textit{13}  & \textit{73} & \textit{11} & \textit{50} \\
\midrule
Pneumonia               & 0.61 & 0.65 & 0.58 & 0.77 & 0.59 & 0.63 \\
AIDS                    & 1.00 & --   & 0.32 & --   & --   & --   \\
Other Cardiovascular    & 0.36 & 0.42 & 0.00 & 0.00 & --   & 0.36 \\
Diarrhea/Dysentery      & 0.51 & 0.50 & 0.36 & 0.25 & 0.43 & 0.69 \\
Road Traffic            & 1.00 & 1.00 & 0.87 & --   & 1.00 & 0.94 \\
Malaria                 & 0.17 & --   & 0.37 & --   & 1.00 & --   \\
Falls                   & 0.75 & 0.80 & 0.00 & 1.00 & 0.25 & 0.33 \\
Drowning                & 0.97 & --   & 0.85 & 0.67 & 0.80 & 0.97 \\
Fires                   & 0.90 & --   & 0.80 & 1.00 & 0.57 & 0.93 \\
Other Defined Causes    & 0.16 & 0.35 & 0.19 & 0.19 & 0.00 & 0.12 \\
\bottomrule
\end{tabular}
\begin{tablenotes}\small
\item Same abbreviations as Table~\ref{tab:cause_specific_adult}. ``--'' indicates no cases.
\end{tablenotes}
\end{threeparttable}
\end{table}

Finally, for neonate, accuracy is strong for \emph{stillbirth}, \emph{preterm}, and \emph{asphyxia}, which have clearer clinical signatures. Performance for neonatal \emph{pneumonia} varies by site and narrative depth.

\begin{table}[ht]
\centering
\caption{Cause-specific Top-1 accuracy (Neonate) by site, with narrative context.}
\label{tab:cause_specific_neonate}
\begin{threeparttable}
\begin{tabular}{lcccccc}
\toprule
\textbf{Cause of Death} & \textbf{AP} & \textbf{Bohl} & \textbf{Dar} & \textbf{Mex} & \textbf{Pmb} & \textbf{UP} \\
\midrule
\textit{Narratives \%} & \textit{64} & \textit{99} & \textit{100} & \textit{75} & \textit{99} & \textit{92} \\
\textit{Avg Words}     & \textit{89} & \textit{170} & \textit{11}  & \textit{75} & \textit{15} & \textit{59} \\
\midrule
Stillbirth              & 0.93 & 0.97 & 0.91 & 0.89 & 0.84 & 1.00 \\
Preterm Delivery        & 0.80 & 0.86 & 0.56 & 0.70 & 0.34 & 0.74 \\
Birth Asphyxia          & 0.50 & 0.80 & 0.60 & 0.42 & 0.53 & 0.56 \\
Congenital Malformations& 0.40 & 0.68 & 0.26 & 0.24 & 0.35 & 1.00 \\
Meningitis/Sepsis       & 1.00 & 0.57 & 0.57 & 0.60 & 1.00 & 0.80 \\
Pneumonia               & 0.00 & 0.19 & 0.05 & --   & 1.00 & 0.15 \\
\bottomrule
\end{tabular}
\begin{tablenotes}\small
\item Same abbreviations as Table~\ref{tab:cause_specific_adult}. ``--'' indicates no cases.
\end{tablenotes}
\end{threeparttable}
\end{table}

\end{document}